\documentclass{article}

\usepackage[final]{corl_2019} 
\usepackage[ruled,vlined]{algorithm2e}
\usepackage{amsmath, amssymb, amsthm}
\usepackage{graphicx}
\usepackage{subcaption}
\usepackage{float}
\usepackage{todonotes}
\usepackage{float}
\usepackage{multicol}
\usepackage{wrapfig}

\graphicspath{{Ressources/}}
\DeclareMathOperator*{\argmax}{arg\,max}

\title{Mastering Rate based Curriculum Learning}

\author{Lucas Willems\thanks{Equal contribution. Code: \url{https://github.com/lcswillems/automatic-curriculum}} \\
\'Ecole Normale Sup\'erieure, Paris \\
\And
Salem Lahlou\footnotemark[1] \\
Mila, Universit\'e de Montr\'eal \\
\And
Yoshua Bengio \\
Mila, Universit\'e de Montr\'eal \\
CIFAR Senior Fellow}

\begin{document}
\maketitle


\begin{abstract}
Recent automatic curriculum learning algorithms, and in particular Teacher-Student algorithms, rely on the notion of learning progress, making the assumption that the good next tasks are the ones on which the learner is making the fastest progress or digress. In this work, we first propose a simpler and improved version of these algorithms. We then argue that the notion of learning progress itself has several shortcomings that lead to a low sample efficiency for the learner. We finally propose a new algorithm, based on the notion of \textit{mastering rate}, that significantly outperforms learning progress-based algorithms.
\end{abstract}



\section{Introduction}

Recently, deep reinforcement learning algorithms have been successfully applied to a wide range of domains (\citep{mnih2015human},~\citep{schulman2017proximal},~\citep{hessel2018rainbow},~\citep{singh2019end}). However, their success relies heavily on dense rewards being given to the agent; and learning in environments with sparse rewards is still a major limitation of RL due to the low sample efficiency of the current algorithms in such scenarios.

In sparse rewards settings, the sample inefficiency is essentially caused by the low likelihood of the agent obtaining a reward by random exploration. Recent attempts to tackle this issue revolve around providing the agent an intrinsic reward that encourages exploring new states of the environment, thus increasing the likelihood of reaching the reward (\citep{ostrovski17a},~\citep{goyal2018transfer},~\citep{kim19a}). An alternative way to improve the sample efficiency is \textit{curriculum learning} (\citep{bengio2009curriculum}). It consists in first training the agent on an easy version of the task at hand, where it can get reward more easily and learn, then training on increasingly difficult versions using the previously learned policy and finally, training on the task at hand. Its usage is not limited to reinforcement learning and robotics tasks, but also to supervised tasks.
Curriculum learning may be decomposed into two parts:
\begin{enumerate}
\item
Defining the \textit{curriculum}, i.e. the set of tasks the learner may be trained on.
\item Defining the \textit{program}, i.e. defining, at each training step, on which tasks to train the learner, given its learning state and the curriculum.
\end{enumerate}

The idea that using a curriculum of increasingly more difficult tasks speeds up neural networks training was put forward in \citep{elman1993learning}. \citep{bengio2009curriculum} paved the way to a wider usage of curriculum learning in the field. \citep{zaremba2014learning} for example used hand-designed curricula to learn to perform memorization and addition with LSTMs (\citep{hochreiter1997long}).~\citep{wu2016training} used curriculum learning to train an actor-critic agent on Doom.~\citep{chevalier-boisvert2018babyai} used small curricula to improve the sample efficiency of ground language learning with imitation learning.~\citep{wein2018} used curriculum learning in the context of training a CNN for visualization tasks. These works rely on hand-designed programs, where there is a performance threshold that allows the learner to advance to the next task, or that increases the number of training examples of the harder tasks in the case of supervised learning. Relying on hand-designed programs creates a significant bottleneck to a broader usage of curriculum learning because:
\begin{itemize}
    \item They are painful to design, requiring a lot of iterations.
    \item They are usually ineffective, leading to learners prone to catastrophic forgetting (\citep{parisi2019continual}).
\end{itemize}

In this work, we focus on \textit{program algorithms} for curriculum learning, i.e. algorithms that decide, given a curriculum and the learning state of the learner, on which tasks to train the learner next.  Several such algorithms emerged recently, based on the notion of \textit{learning progress} (\citep{matiisen2017teacher},~\citep{graves2017automated},~\citep{fournier2018}). \citep{matiisen2017teacher} proposed four algorithms (called Teacher-Student) based on the assumption that the good next tasks are the ones on which the learner is making the fastest progress or digress. Our contributions are two-fold:
\begin{enumerate}
    \item We propose a new Teacher-Student algorithm that is simpler to use, because it has one less hyperparameter to tune, and we show it has improved performances in terms of stability and sample complexity. 
    \item Based on observed shortcomings of the Teacher-Student algorithms where the learner may be mainly trained on tasks it \textit{cannot learn yet} or tasks it \textit{already learned}, we make the assumption that the good next tasks are the ones that are \textit{learnable but not learned yet}, and we thus introduce a new algorithm based on the notion of \textit{mastering rate}. We provide experimental evidence that our algorithm significantly outperforms the Teacher-Student algorithms on various Reinforcement Learning and Supervised Learning tasks.
\end{enumerate}

Our mastering rate based algorithm requires the input curriculum to contain more structure than those required by the Teacher-Student algorithms. Essentially, the different tasks should be ordered by difficulty. However, this doesn't create a significant overhead given that, to the best of our knowledge, all curricula tackled in the recent literature present a natural ordering of the tasks (e.g. size of the environment, number of objects in the environment,...).


\section{Background}
\label{section2}

\subsection{Curriculum learning}

Curriculum learning can be decomposed into:
\begin{enumerate}
\item Defining the \textit{curriculum}, i.e. the set of tasks the learner may be trained on. \\
Formally, a \textit{curriculum} \(\mathcal{C}\) is a set of tasks \(\{c_1, ..., c_n\}\). A \textit{task} $c$ is a set of environments (in RL) or examples (in supervised learning) of similar type, with a sampling distribution.
\item Defining the \textit{program}, i.e. defining, at each training step, on which tasks to train the learner, given its learning state and the curriculum. \\
Formally, a \textit{program} \footnotemark \(\ \textbf{d}:\mathbb{N}\rightarrow\mathcal{D}^\mathcal{C}\) is a time-varying sequence of distributions over \(\mathcal{C}\).
\footnotetext{What we call ``program" is called ``syllabus'' in \citep{graves2017automated}.}
\end{enumerate}

To unify the program algorithms introduced in  \citep{matiisen2017teacher}, \citep{graves2017automated} and \citep{fournier2018}, let's observe that defining a program can be decomposed into:
\begin{enumerate}
    \item Defining an \textit{attention program} \(\textbf{a}:\mathbb{N}\rightarrow \mathcal{A}^\mathcal{C}\), i.e. a time-varying sequence of attentions over tasks \(\mathcal{C}\). An \textit{attention \(a\) over \(\mathcal{C}\)} is a family of non-negative numbers indexed by \(\mathcal{C}\), i.e. \(a = (a_{c})_{c\in\mathcal{C}}\) with \(a_c \geq 0\).
    \item Defining an \textit{attention-to-distribution converter} (called \textit{A2D converter}) \(\Delta:\mathcal{A}^\mathcal{C}\rightarrow\mathcal{D}^\mathcal{C}\) that converts an attention over \(\mathcal{C}\) into a distribution over \(\mathcal{C}\).
\end{enumerate}

Hence, a program \(\textbf{d}:\mathbb{N}\rightarrow\mathcal{D}^\mathcal{C}\) can be seen as the composition of an attention program \(\textbf{a}:\mathbb{N}\rightarrow \mathcal{A}^\mathcal{C}\) and an A2D converter \(\Delta:\mathcal{A}^\mathcal{C}\rightarrow\mathcal{D}^\mathcal{C}\), i.e. \(\textbf{d}(t) = \Delta(\textbf{a}(t))\).

Thanks to this observation, each program algorithm in \citep{matiisen2017teacher}, \citep{graves2017automated} and \citep{fournier2018} is a particular case of algorithm \ref{algo:rl} with a specific implementation of \(\Delta\) and \(\textbf{a}\).

\begin{center}
\begin{algorithm}[H]
\label{algo:rl}
\caption{Generic program algorithm (RL version)}

\SetKwInOut{Input}{input}
\Input{A curriculum \(\mathcal{C}\) \; \\
       A learner \(\mathbf{A}\);}
\For{$t\gets1$ \KwTo $T$}{
    Compute \(\textbf{a}(t)\) \;
    Deduce \(\textbf{d}(t):=\Delta(\textbf{a}(t))\) \;
    Draw task \(c\) from \(\textbf{d}(t)\) and environment \(e\) from \(c\) \;
    Train \(\mathbf{A}\) on \(e\) and observe return \(r_{t}^{c}\);
}
\end{algorithm}
\end{center}

For some RL algorithms, a parallelized version of algorithm \ref{algo:rl} might speed up learning: see algorithm \ref{algo:rl-parallel} in appendix \ref{appendix:algos} for such a version. Moreover, in supervised learning settings, a batch version of algorithm \ref{algo:rl} might be more appropriate: see algorithm \ref{algo:sl} in appendix \ref{appendix:algos} for such a version.

\subsection{Teacher-Student program algorithms}

Two A2D converters were introduced in \citep{matiisen2017teacher}:
\begin{itemize}
\item the \textit{greedy argmax} converter (called \textit{gAmax}): \(\Delta^{gAmax}(a) := (1-\varepsilon)\Delta^{Amax}(a) + \varepsilon \cdot u\) \\
with \; \(\Delta^{Amax}(a)_c = \begin{cases}1 & \text{if \(c=\argmax_{c'} a_{c'}\)} \\
0 & \text{otherwise} \\
\end{cases}\),\; \(\varepsilon \in [0, 1]\) and \(u\) the uniform distribution.
\item the \textit{Boltzmann} converter: \(\Delta^{Boltz}(a)_c := \frac{\exp(a_c/\tau)}{\sum_{c'}\exp(a_{c'}/\tau)}\).
\end{itemize}
and four attention program algorithms: \textit{Online}, \textit{Naive}, \textit{Window} and \textit{Sampling}.

All their attention program algorithms are based on the idea that the attention given to task $c$ must be the absolute value of the learning progress on it, i.e. \(\textbf{a}_c(t) := |\beta_c(t)|\) where \(\beta_{c}(t)\) is an estimate of the \textit{learning progress of the learner} \(\mathbf{A}\) on \(c\). By doing so, the teacher (task chooser) encourages the student (learner) to train on tasks on which it is making the fastest progress or digress.

Each attention program algorithm differs by its learning progress estimate \(\beta_{c}(t)\). For example, the Window algorithm makes use of:
\begin{enumerate}
    \item \(\beta^{Linreg}_c := \text{slope of the linear regression of } (t_1, r_{t_1}^c), ..., (t_K, r_{t_K}^c)\) \\
    with  \(t_1, ..., t_K\) the \(K\) last time-steps the learner was trained on a sample of \(c\) and \(r_{t_1}^c, ..., r_{t_K}^c\) the performances it obtained at these time-steps.
    \item \(\beta_{c}(t) := \alpha \beta^{Linreg}_{c}(t) + (1-\alpha)\beta_{c}(t-1)\).
\end{enumerate}


\section{A simpler and improved Teacher-Student algorithm}
\label{section3}

The \textit{gAmax Window} program algorithm is the one obtaining the best performances in \citep{matiisen2017teacher}. In this section, we propose a simpler version, requiring one less hyperparameter to tune, and showing improved performances. It consists in:
\begin{itemize}
    \item Removing the weighted moving average of the Window program algorithm, because the hyperparameter $K$ of the linear regression already makes it capture enough information from the past (see experiment results in section \ref{23}), making the moving average redundant.\\
    This new attention program, called \textit{Linreg}, uses: \(\beta_c := \beta^{Linreg}_c\) as a learning progress estimator, and has no hyperparameter $\alpha$ to tune. 
    \item Replacing the \textit{gAmax} A2D converter with the \textit{greedy proportional} converter (called \textit{gProp}):\\ \(\Delta^{gProp}(a) := (1-\varepsilon)\cdot \Delta^{Prop}(a) + \varepsilon \cdot u\) \; with \; \(\Delta^{Prop}(a)_c := \frac{a_c}{\sum_{c'} a_{c'}}\).\\
    It makes learning more stable (see experiment results in section \ref{23}) because, if two tasks $A$ and $B$ have almost equal attentions (i.e. almost equal learning progress here), they will get almost equal training probabilities with \textit{gProp}, whereas, with \textit{gAmax}, one will get a low probability of $\varepsilon/2$ and the other a high probability of $1-\varepsilon/2$.
\end{itemize}

\paragraph{Naming convention.} Hereinafter, we will refer to the Teacher-Student algorithm using gProp and Linreg as the \textbf{gProp Linreg} algorithm. 

In section \ref{23}, we make experiments demonstrating the performances of the gProp Linreg algorithm, compared to other Teacher-Student algorithms.


\section{A Mastering Rate based algorithm}
\label{section4}

\label{mr-algorithm}
In this section, and for the sake of illustration, let's consider the curriculum ${\mathcal{C} = (A, B)}$ where, for a given learner, task $A$ requires a large amount of experience (training examples in supervised learning or environment interactions in reinforcement learning), and task $B$ requires even more experience, but learning to accomplish task $A$ makes task $B$ easier. 

Learning progress-based Teacher-Student algorithms introduced by \citep{matiisen2017teacher} rely on the assumption that the good next tasks are the ones \textit{on which the learner is making the fastest progress or digress}. However, two highly sample inefficient situations may occur:
\begin{enumerate}
\item The learner may be mainly trained on tasks it \textbf{already learned}. On $\mathcal{C}$, once the learner reaches an almost perfect performance on task $A$, it will continue to have a non-zero absolute learning progress on $A$ and zero learning progress on $B$. It will thus continue to be mainly trained on $A$ even if it has nearly perfectly learned it, unless the performance on $A$ plateaues. This situation is not desirable given that, in most curricula, switching to task $B$ is likely to improve the performances on $A$ as well. This scenario is highlighted in frame B of figure \ref{fig:BUP-ReturnProba} in appendix \ref{appendix:shortocmings}.
\item The leaner may be mainly trained on tasks it \textbf{can't learn yet}. On $\mathcal{C}$, during the initial training period where the learner obtains zero performance on both tasks, the learning progress on $A$ and $B$ will be zero and the tasks will be equally sampled. Hence, half the time, the learner will be trained on $B$, which it can't start learning without first making some progress on $A$. This worsens as the number of tasks in the curriculum increases. This scenario is highlighted in frame A of figure \ref{fig:BUP-ReturnProba} in appendix \ref{appendix:shortocmings}.
\end{enumerate}

To prevent these two highly sample inefficient situations from occurring, we introduce a new algorithm based on the assumption that the good next tasks to train on are the ones \textit{that are learnable but not learned yet}. The central notion is not learning progress anymore, but \textit{mastering rate}. It presupposes we have access to the underlying structural relationships between the different tasks.

In the remainder of this section, we will first define what are learned and learnable tasks and then, introduce the new algorithm and the intuitions behind.

\subsection{Learned \& learnable tasks}\label{sec:learned-learnable}
\paragraph{Learned tasks.}
In supervised learning, we can say that a certain task \textit{is learned} by the learner if its test accuracy (or another performance metric that depends on the task) is close to one. But this definition of learned task doesn't hold in RL where the maximum (or minimum) possible return may vary among different environment instances of the same task or even not be known a priori. Hence, we provide hereafter a generic definition of \textit{learned task}, applicable in reinforcement or supervised learning settings (where \textit{return} means test accuracy or whatever other performance measure used).


First, let's define a \textit{min-max curriculum} as a curriculum \(\mathcal{C}=\{c_1, ..., c_n\}\) along with a family \((\hat{m}_{c})_{c\in\mathcal{C}}\) (resp. \((\hat{M}_{c})_{c\in\mathcal{C}}\)) where  \(\hat{m}_c\) (resp. \(\hat{M}_c\)) is an estimate of the minimum (resp. maximum) possible mean return the learner can get on task \(c\).
If the estimation is not perfect, it has to be higher (resp. lower) than the true minimum (resp. maximum) mean return.
These estimaties must be given by the practitioner when defining the min-max curriculum.

On such a curriculum, let's define, for a task \(c\), the \textit{running  mean return} \(\bar{r}_c(t)\), the \textit{running minimum mean return} $\bar{m}_c(t)$ and the \textit{running maximum mean return} $\bar{M}_c(t)$:
$$\begin{cases}\bar{r}_c(0) := \hat{m}_{c} &  \\
\bar{r}_c(t) := \frac{r_{t_1}^c+...+r_{t_K}^c}{K}   \\
\end{cases}
,\quad
\begin{cases}
\bar{m}_c(0) := \hat{m}_c & \\
\bar{m}_c(t):=\min(\bar{r}_c(t), \bar{m}_c(t-1)) 
\end{cases}
,\quad
\begin{cases}
\bar{M}_c(0) := \hat{M}_c & \\
\bar{M}_c(t):=\max(\bar{r}_c(t), \bar{M}_c(t-1)) 
\end{cases}
$$
where \(t_{1}, ..., t_K\) are the \(K\) last time-steps the learner was trained on a sample of \(c\) and where \(r_{t_i}^c\) is the return obtained by the learner at these time-steps. At each time-step, \(\bar{m}_c\) (resp. \(\bar{M}_c\)) is a better estimate of the minimum (resp. maximum) possible mean return of task \(c\).

From this, we can define the \textit{mastering rate} of a task \(c\) as:
$$\mathcal{M}_c(t) :=\frac{\bar{r}_c(t) - \bar{m}_c(t)}{\bar{M}_c(t) - \bar{m}_c(t)}.$$

Intuitively, a task \(c\) is said to be:
\begin{itemize}
\item ``learned" if \(\mathcal{M}_c(t)\) is near 1, because \(\bar{r}_c(t)\) would be near \(\bar{M}_c(t)\)
\item ``not learned" if  \(\mathcal{M}_c(t)\) is near 0, because \(\bar{r}_c(t)\) would be near \(\bar{m}_c(t)\)
\end{itemize}

\paragraph{Learnable tasks.}

An \textit{ordered curriculum} \(\mathcal{O}^{\mathcal{C}}\) is an directed acyclic graph over tasks \(\mathcal{C}\). An edge goes from task \(A\) to task \(B\) if it is preferable to learn \(A\) before \(B\). A \textit{min-max ordered curriculum} is an ordered curriculum along with a minimum and maximum mean return estimate for each task. On such a curriculum, we can define, for a task \(c\), the \textit{learnability rate}:
$$\mathcal{M}_{\text{Anc }c}(t):= \min_{c' |c'\rightsquigarrow c} \mathcal{M}_{c'}(t)$$
where \(c'\rightsquigarrow c\) means that \(c'\) is an ancestor of \(c\) in the graph \(\mathcal{O}^{\mathcal{C}}\). If \(c\) has no ancestor, \(\mathcal{M}_{\text{Anc }c}(t)=1\). Intuitively, a task \(c\) is said to be:
\begin{itemize}
\item ``learnable'' if \(\mathcal{M}_{\text{Anc }c}(t)\) is near 1, because the mastering rate of all the ancestor tasks would be near 1, meaning they would all be mastered.
\item ``not learnable'' if \(\mathcal{M}_{\text{Anc }c}(t)\) it is near 0, because the mastering rate of at least one of the ancestor tasks would
be near 0, meaning it would not be mastered.
\end{itemize}

\subsection{A mastering rate based algorithm (MR algorithm)}\label{sec:mr-alg}


Unlike Teacher-Student algorithms that require a curriculum as input, the mastering rate based algorithm (MR algorithm) requires a min-max ordered curriculum. However, this doesn't create a significant overhead: for all the experiments in \citep{matiisen2017teacher},~\citep{graves2017automated}, and~\citep{fournier2018} for example, min-max ordered curricula could have been provided easily.

\paragraph{Attention program.} Let's define how attention is given to each task $c$:
\begin{equation}
\label{eq:mr-attention}
a_c(t) =\left[\left(\mathcal{M}_{\text{Anc }c}(t)\right)^p\right]\left[\delta(1-\mathcal{M}_{c}(t)) + (1-\delta)|\hat{\beta}^{Linreg}_c(t)| \right]\left[1-\mathcal{M}_{\text{Succ } c}(t)\right]
\end{equation}
where:
\begin{itemize}
\item \(\hat{\beta}^{Linreg}_c(t):= \frac{\beta^{Linreg}_c(t)}{\max_{c'}|\beta^{Linreg}_{c'}(t)|}\) if possible.\  Otherwise, \(\hat{\beta}^{Linreg}_c(t):=0\);
\item
\(\mathcal{M}_{\text{Succ }c}(t):= \min_{c'|c\rightarrow c'} \mathcal{M}_{c'}(t)\). If \(c\) has no successor, \(\mathcal{M}_{\text{Succ }c}(t):=0\).
\end{itemize}

The attention \(a_c(t)\) given to a task \(c\) is the product of three terms (written inside square brackets):
\begin{itemize}
\item The first term gives attention only to tasks that are learnable (i.e. $\mathcal{M}_{\text{Anc }c}(t)$ close to one) . The power \(p\) controls how much a task should be learnable in order to get attention.
\item The second term,  when \(\delta=1\), gives attention only to tasks that are not learned yet (i.e. $\mathcal{M}_{c}(t)$ close to zero).
When \(\delta \neq 1\), it also gives attention to tasks that are progressing or regressing quickly. Using \(\delta \neq 1\) prevents situations where the learner stops training on task $c$ as soon as $\bar{r}_c(t)$ reaches $\bar{M}_c(t)$ whereas it is still making progress on it, because $\bar{M}_c(t)$ might be a wrong estimate of the maximum possible mean return.
\item The last term gives attention only to tasks whose successors are not learned yet. Without it, learned tasks whose successors are learned might still have attention because of \(\hat{\beta}^{Linreg}_c(t)\) in the second term.
\end{itemize}
In order to ensure that easier tasks are sampled regularly in the first stages of training, each task first yields a part \(\gamma_{pred}\) of its attention to its predecessors, defining a new attention $a'$:
\begin{equation}\label{eq:MR-pred-rd}
a_c'(t) := (1-\gamma_{pred})a_c(t)+\sum_{c' | c\rightarrow c'} \frac{\gamma_{pred}}{n_{\rightarrow c'}}a'_{c'}(t)
\end{equation}
where \(n_{\rightarrow c}\) is the number of predecessors of $c$. Then, each task yields a part \(\gamma_{succ}\) of its attention to its successors:
\begin{equation}\label{eq:MR-succ-rd}
a_c{''}(t) := (1-\gamma_{succ})a^{'}_c(t)+\sum_{c' | c'\rightarrow c} \frac{\gamma_{succ}}{n_{c'\rightarrow}}a'_{c'}(t)
\end{equation}
where \(n_{c\rightarrow}\) is the number of successors of $c$.

\paragraph{A2D converter.} If we don't want to sample learned or not learnable tasks, we shouldn't use the gAmax or gProp A2D converters defined in sections \ref{section2} and \ref{section3}. We prefer the Prop A2D converter in this scenario.

Algorithm \ref{alg:MR-algorithm} summarizes the MR algorithm for the RL setting. For supervised learning, a batch version could be obtained by easily adapting \ref{algo:sl} in appendix \ref{appendix:algos}.

\begin{center}
\begin{algorithm}[H]
\label{alg:MR-algorithm}
\caption{MR algorithm}

\SetKwInOut{Input}{input}
\Input{A min-max ordered curriculum \(\mathcal{C}\) \; \\
       A learner \(\mathbf{A}\);}

$\beta^{Linreg} := 0$ \;
\For{$t\gets1$ \KwTo $T$}{
    Compute attention \(a\) using equation \ref{eq:mr-attention} \;
    Compute redistributed attention \(a''\) following equation \ref{eq:MR-pred-rd} and \ref{eq:MR-succ-rd}\;
    \(d:=\Delta^{Prop}(a'')\) \;
    Draw task \(c\) from \(d\) and environment \(e\) from \(c\) \;
    Train \(\mathbf{A}\) on \(e\) and observe return \(r_{t}^c\) \;
    \(\beta^{Linreg}_c := \text{slope of lin. reg. of } (t_1, r_{t_1}^c), ..., (t_K, r_{t_K}^c)\) \;
}
\end{algorithm}
\end{center}

Finally, we can remark that the MR\ algorithm is a generalization of Teacher-Student algorithms. If we consider min-max ordered curricula without edges (i.e. just curricula),  if \(\delta=0\), and if we use the gProp dist converter instead of the Prop one, then the MR algorithm is exactly the gProp Linreg algorithm. Note that the MR algorithm can be adpated to use different learning progress estimators $\hat{\beta}$ in equation \ref{eq:mr-attention}, and different A2D converters.

\section{Experimental setting}
\label{section5}
In this section we will present the experimental setting we used to demonstrate our results. 

For supervised learning, we evaluate our algorithms on the task of adding two 9-digit decimal numbers with LSTMs. The learner (a neural network) receives as input two numbers separated by the plus sign, and outputs the string corresponding to the sum of those two numbers.~\citep{zaremba2014learning} used a curriculum of 9 tasks of increasing difficulty (that differ by the number of digits of each input number) to learn to perform the task in reasonable time. The batch version of the algorithms (e.g. algorithm \ref{algo:sl}) is used for this task.

For reinforcement learning, three curricula were used to evaluate the algorithms:
\begin{enumerate}
\item the \textit{BlockedUnlockPickup} curriculum: a sequence of 3 tasks of increasing difficulty. See figure \ref{fig:blockedunlockpickup} in appendix \ref{app:curricula} for snapshots and more details.
\item the \textit{KeyCorridor} curriculum: a sequence of 6 tasks of increasing difficulty. See figure \ref{fig:keycorridor} in appendix \ref{app:curricula} for snapshots and more details.
\item the \textit{ObstructedMaze} curriculum, made of 6 tasks. However, they can't be seen as a sequence of tasks of increasing difficulty because some tasks are not harder or easier than others: they require different abilities. See figure \ref{fig:obstructedmaze} in appendix \ref{app:curricula} for snapshots and more details.
\end{enumerate}

These tasks come from \href{https://github.com/maximecb/gym-minigrid}{Gym MiniGrid} environments (\citep{gym_minigrid}).\ They are partially observable: the observations use a compact and efficient encoding, with just 3 input values per visible grid cell, making the observation shape \(7\times7\times3\) at each time-step. The gray-shaded areas in the snapshots of appendix \ref{app:curricula} represent this partial view.

In all our RL experiments, the agent is trained using Proximal Policy Optimization (\citep{schulman2017proximal}). We used a convolutional network to encode the observations, along with an LSTM to handle the partial observability of the environment. The state embedding is then fed to a policy and value networks.

The environments have sparse rewards: the agent gets a final reward of \(1-\frac{n}{n_{max}}\) when the instruction is executed in \(n\) steps with \(n\leq n_{max}\) (\(n_{max}\) depends on the environment); otherwise, it gets a reward of \(0\). Note that different instances of the same task (e.g. the ``Unlock'' task) have different maximum possible rewards, given that the agent is initially placed randomly in the environment.

\section{Results}

\subsection{Improved Teacher-Student algorithm}
\label{23}
\begin{figure}
     \centering
     \begin{subfigure}[b]{0.45\textwidth}
         \centering
         \includegraphics[width=\textwidth]{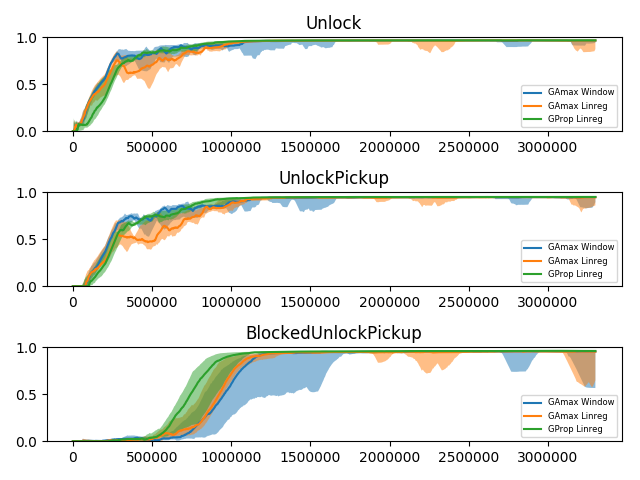}
         \caption{BlockedUnlockPickup curriculum}
         \label{fig45}
     \end{subfigure}
     \hfill
     \begin{subfigure}[b]{0.45\textwidth}
         \centering
         \includegraphics[width=\textwidth]{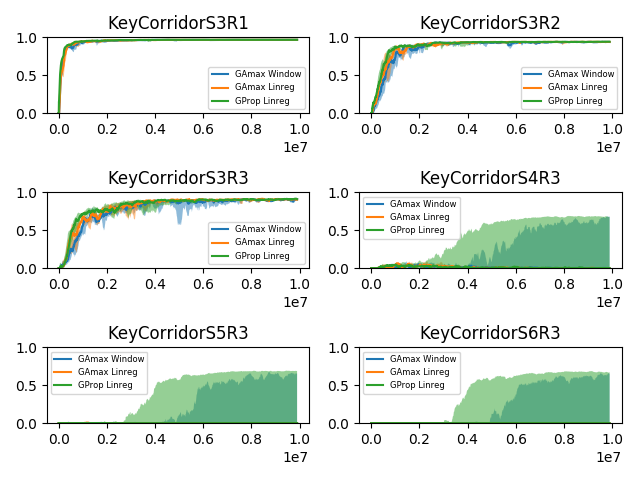}
         \caption{KeyCorridor curriculum}
         \label{fig46}
     \end{subfigure}
        \caption{gAmax\ Window, gAmax Linreg and gProp Linreg algorithms were each tested with 10 different agents (seeds) on the KeyCorridor and the BlockedUnlockPickup curriculum.\  The median return during training, along with a confidence interval representing the first and last quartiles, are plotted. The x-axis represents the number of frames.}
        \label{fig:two graphs}
\end{figure}

In this section, we show the results of the Teacher-Student algorithm proposed in section \ref{section3}, that uses the Linreg attention program and the gProp A2D converter (gProp Linreg). We trained different agents on two Minigrid curricula. The results are depicted in figures \ref{fig45} and \ref{fig46}.

Figures \ref{fig45} and \ref{fig46} justify our usage of the Linreg attention program and the gProp A2D converter:
\begin{itemize}
\item the gAmax Linreg algorithm has comparable performances to the gAmax Window algorithm, as asserted in section \ref{section3}. It even seems more stable because the gap between the first and last quartiles is smaller.
\item the gProp Linreg algorithm performs better than the gAmax Linreg and gAmax Window algorithm, as asserted in section \ref{section3}.
\end{itemize}
Given the success of the gProp Linreg algorithm for these curricula, it will be the only one used to compare Teacher-Student algorithms to the MR algorithm in section \ref{section54}.

\subsection{Decimal number addition curriculum}
\label{section53}
\begin{figure}
\centering
\includegraphics[width=.5\linewidth]{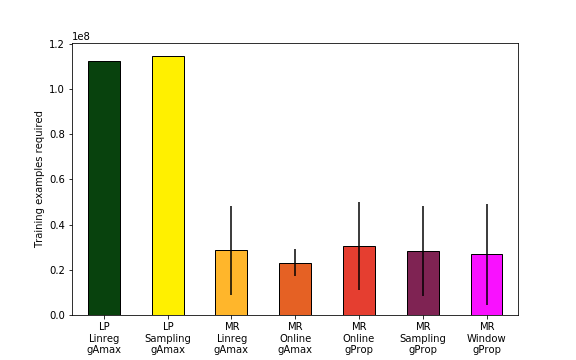}
\caption{Sample complexity of the decimal number addition curriculum (lower is better). }
\label{fig:LSTM}
\end{figure}

We trained an LSTM on the addition curriculum defined in section \ref{section5}. We tried different learning progress estimators and A2D converters for the MR algorithm, and we compared their performances to two of the best performing ones (Linreg gAmax and Sampling gAmax) used in~\citep{matiisen2017teacher} for this task. Similar to~\citep{matiisen2017teacher}, we used an LSTM with 128 units for both the encoder and the decoder, and we passed the last output of the encoder to all inputs of the decoder. Figure \ref{fig:LSTM} shows, for the configurations we tried, the number of training examples required to reach $99\%$ validation accuracy. We observe that regardless of the attention program and the A2D converter, the MR algorithm significantly outperforms the Teacher-Student algorithms (denoted by ``LP'' in the figure) for this task. We used four different random seeds for the MR algorithm. The hyperparameters we used can be found in appendix \ref{app:hyperparam}.

\subsection{Gym Minigrid curricula}
\label{section54}
As mentioned in section \ref{section5}, we trained a PPO agent on the three Minigrid curricula. We used 10 seeds for statistical significance. The results are depicted in figure \ref{fig:555} for the curricula with 6 tasks (KeyCorridor and ObstructedMaze), and in figure \ref{fig:bupmr} in appendix \ref{app:blockedunlockpickup} for the curriculum with 3 tasks (BlockedUnlockPickup).
The MR algorithm with \(\delta = 0.6\)  (see appendix \ref{app:ordered-curricula} for the min-max ordered curricula given to this algorithm) outperforms the gProp Linreg algorithm on the three Minigird curricula, especially on:
\begin{itemize}
\item
the KeyCorridor curriculum (see figure \ref{fig55}) where the median return of the gProp Linreg algorithm is near 0 after 10M time-steps on S4R3, S5R3 and S6R3 while the first quartile of the MR\ algorithm is higher than 0.8 after 6M time-steps.
\item the ObstructedMaze curriculum (see figure \ref{fig56}) where the last quartile of the gProp Linreg algorithm is near 0 after 10M time-steps on all the tasks while the last quartile of the MR\ algorithm is higher than 0.7 after 5M time-steps on 1Dl, 1Dlh, 1Dlhb, 2Dl, 2Dlh.
\end{itemize}

\begin{figure}
     \centering
     \begin{subfigure}[b]{0.48\textwidth}
         \centering
         \includegraphics[width=\textwidth]{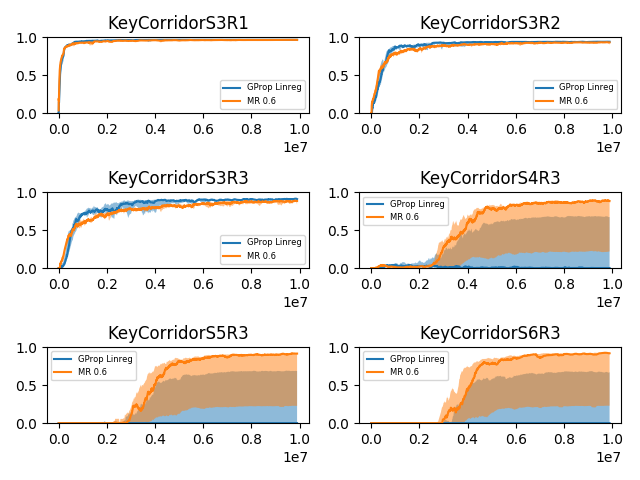}
         \caption{KeyCorridor curriculum}
         \label{fig55}
     \end{subfigure}
     \hfill
     \begin{subfigure}[b]{0.48\textwidth}
         \centering
         \includegraphics[width=\textwidth]{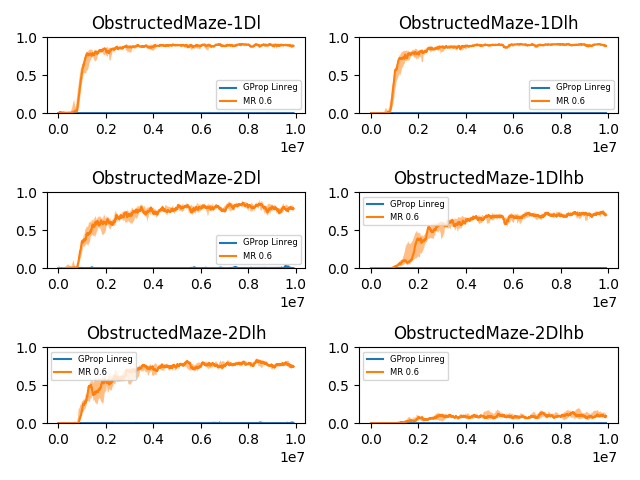}
         \caption{ObstructedMaze curriculum}
         \label{fig56}
     \end{subfigure}
        \caption{gProp Linreg and MR with \(\delta = 0.6\)  were each tested with 10 different agents (seeds) on the ObstructedMaze and KeyCorridor curricula.\  The median return during training, along with a confidence interval representing the first and last quartiles, are plotted. The x-axis represents the number of frames. Note that the agent trained with the gProp Linreg algorithm is not able to obtain a positive return in any of the ObstructedMaze curriculum tasks.}
        \label{fig:555}
\end{figure}


\section{Conclusion}
\label{sec:conclusion}
We experimentally showed that learning progress based program algorithms have different shortcomings, and proposed a new algorithm, based on the notion of mastering rate, that addresses them. While it requires more input from the practitioner (a min-max ordered curriculum), it remains a small cost for significant gains in performance: learning is much more sample efficient and robust.



\clearpage
\acknowledgments{We thank Jacob Leygonie and Claire Lasserre for fruitful conversations regarding these ideas, Victor Schmidt for useful remarks, and Compute Canada for computing support.}


\bibliography{main}  
\clearpage

\appendix

\section{Alternative versions of algorithm \ref{algo:rl}}
\label{appendix:algos}

\begin{center}
\begin{algorithm}[H]
\label{algo:rl-parallel}
\caption{Generic program algorithm (parallelized RL version)}

\SetKwInOut{Input}{input}
\Input{A curriculum \(\mathcal{C}\) \; \\
       A learner \(\mathbf{A}\);}
\For{$t\gets1$ \KwTo $T$}{
    Compute \(\textbf{a}(t)\) \;
    Deduce \(\textbf{d}(t):=\Delta(\textbf{a}(t))\) \;
    Draw tasks \(c_{i_1}, ..., c_{i_k}\) from \(\textbf{d}(t)\) and environments \(e_{i_1}, ..., e_{i_k}\) from \(c_{i_1}, ..., c_{i_k}\) \;
    Train \(\mathbf{A}\) on \(e_{i_1}, ..., e_{i_k}\) and observe returns \(r_{t}^{c_{i_1}},\dots,r_{t}^{c_{i_k}}\) \;
}
\end{algorithm}
\end{center}

\begin{center}
\begin{algorithm}[H]
\label{algo:sl}
\caption{Generic program algorithm (Supervised learning version)}

\SetKwInOut{Input}{input}
\Input{A curriculum \(\mathcal{C}\) \; \\
       A learner \(\mathbf{A}\);}
\For{$t\gets1$ \KwTo $T$}{
    Compute \(\textbf{a}(t)\) \;
    Deduce \(\textbf{d}(t):=\Delta(\textbf{a}(t))\) \;
    Draw a minibatch \(\mathcal{B}\) of $k$ examples according to \(\textbf{d}(t)\) \;
    Train \(\mathbf{A}\) on \(\mathcal{B}\) and observe test scores (or another performance measure) \(r_{t}^{c_1},\dots,r_{t}^{c_n}\) \;
}
\end{algorithm}
\end{center}

\section{Curricula}\label{app:curricula}

In Gym MiniGrid, the action space is discrete: \{go forward, turn left, turn right, toggle (i.e. open or close), pick-up, drop\}.

\begin{figure}[H]
  \begin{subfigure}[b]{0.32\linewidth}
    \centering
    \includegraphics[width=\linewidth]{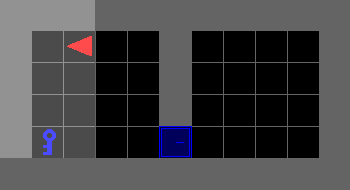}
    \caption{\textit{Unlock}.
    \\ \(n_{max}=288 \)}
  \end{subfigure}
  \hfill
  \begin{subfigure}[b]{0.32\linewidth}
    \centering
    \includegraphics[width=\linewidth]{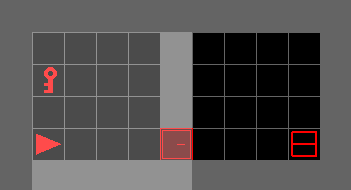}
    \caption{\textit{UnlockPickup}. \\
    \(n_{max}=288\)}
  \end{subfigure}
  \hfill
  \begin{subfigure}[b]{0.32\linewidth}
    \centering
    \includegraphics[width=\linewidth]{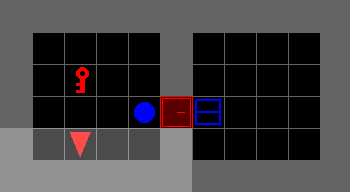}
    \caption{\textit{BlockedUnlockPickup}. \\
    \(n_{max}=576\)}
  \end{subfigure}
  \caption{\textit{BlockedUnlockPickup} curriculum. In Unlock, the agent has to open the locked door. In the others, it has to pick up the box. In UnlockPickup, the door is locked and, in BlockedUnlockPickup, it is locked and blocked by a ball. The position and color of the door and the box are random.}
\label{fig:blockedunlockpickup}
\end{figure}

\begin{figure}[H]
  \begin{subfigure}[b]{0.32\linewidth}
    \centering
    \includegraphics[width=\linewidth]{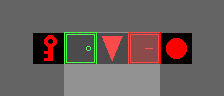}
    \caption{\textit{S3R1}. \\ \(n_{max}=270\)}
  \end{subfigure}
  \hfill
  \begin{subfigure}[b]{0.32\linewidth}
    \centering
    \includegraphics[width=\linewidth]{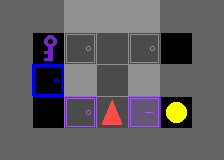}
    \caption{\textit{S3R2}. \\
    \(n_{max}=270\)}
  \end{subfigure}
  \hfill
  \begin{subfigure}[b]{0.32\linewidth}
    \centering
    \includegraphics[width=\linewidth]{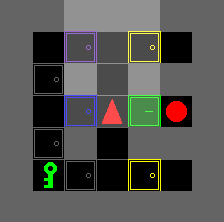}
    \caption{\textit{S3R3}. \\
    \(n_{max}=270\)}
  \end{subfigure}
  \\ \\
  \begin{subfigure}[b]{0.32\linewidth}
    \centering
    \includegraphics[width=\linewidth]{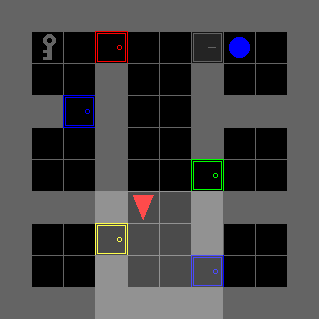}
    \caption{\textit{S4R3}.
    \\ \(n_{max}=480 \)}
  \end{subfigure}
  \hfill
  \begin{subfigure}[b]{0.32\linewidth}
    \centering
    \includegraphics[width=\linewidth]{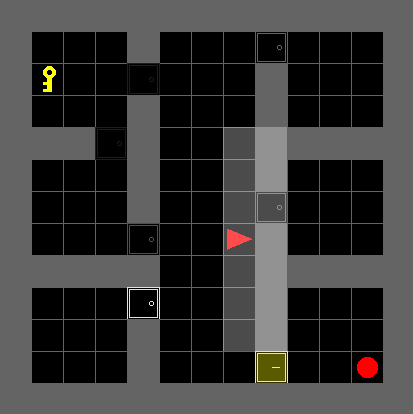}
    \caption{\textit{S5R3}. \\
    \(n_{max}=750\)}
  \end{subfigure}
  \hfill
  \begin{subfigure}[b]{0.32\linewidth}
    \centering
    \includegraphics[width=\linewidth]{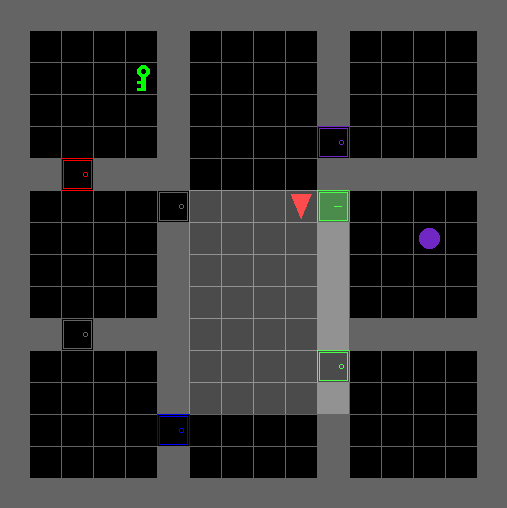}
    \caption{\textit{S6R3}. \\
    \(n_{max}=1080\)}
  \end{subfigure}
  \caption{\textit{KeyCorridor} curriculum. The agent has to pick up the ball.\ However, the ball is in a locked room and the key in another room. The number of rooms and their size gradually increase. The position and color of the key and the doors are random.}
\label{fig:keycorridor}
\end{figure}

\begin{figure}[H]
  \begin{subfigure}[b]{0.3\linewidth}
  \begin{subfigure}[b]{\linewidth}
    \centering
    \includegraphics[width=\linewidth]{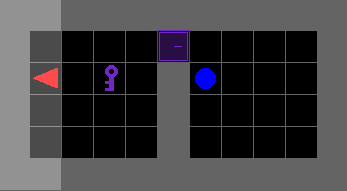}
    \caption{\textit{1Dl}.
    \\ \(n_{max}=288 \)}
  \end{subfigure} \\\\
  \begin{subfigure}[b]{\linewidth}
    \centering
    \includegraphics[width=\linewidth]{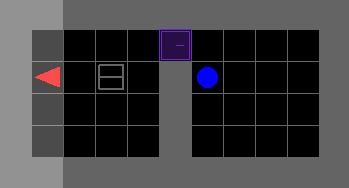}
    \caption{\textit{1Dlh}. \\
    \(n_{max}=288\)}
  \end{subfigure} \\\\
  \begin{subfigure}[b]{\linewidth}
    \centering
    \includegraphics[width=\linewidth]{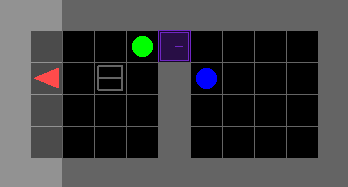}
    \caption{\textit{1Dlhb}. \\
    \(n_{max}=288\)}
  \end{subfigure}
  \end{subfigure}
  \hfill
  \begin{subfigure}[b]{0.6\linewidth}
  \begin{subfigure}[b]{0.32\linewidth}
    \centering
    \includegraphics[width=\linewidth]{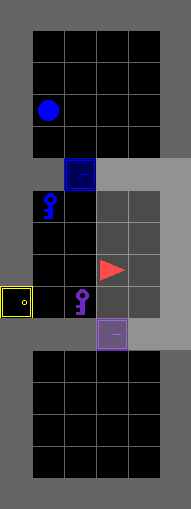}
    \caption{\textit{2Dl}.
    \\ \(n_{max}=576 \)}
  \end{subfigure}
  \hfill
  \begin{subfigure}[b]{0.32\linewidth}
    \centering
    \includegraphics[width=\linewidth]{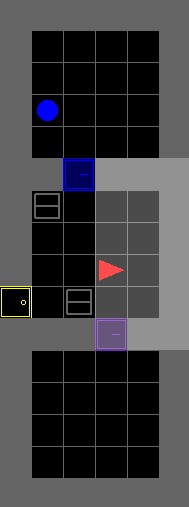}
    \caption{\textit{2Dlh}. \\
    \(n_{max}=576\)}
  \end{subfigure}
  \hfill
  \begin{subfigure}[b]{0.32\linewidth}
    \centering
    \includegraphics[width=\linewidth]{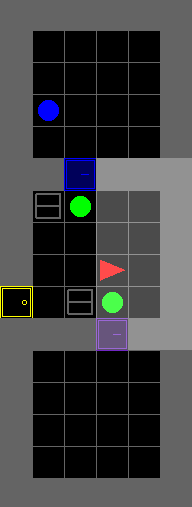}
    \caption{\textit{2Dlhb}. \\
    \(n_{max}=576\)}
  \end{subfigure}
  \end{subfigure}
  \caption{\textit{ObstructedMaze} curriculum. The agent has to pick up the blue ball. The boxes hide a key. A key can only open a door of its color. The number of rooms and the difficulty of opening doors increases.}
\label{fig:obstructedmaze}
\end{figure}

Three curricula were used to evaluate the algorithms in the RL setting: BlockedUnlockPickup (3 tasks), KeyCorridor (6 tasks) and ObstructedMaze (6 tasks). Snapshots of the corresponding environments are shown in figures \ref{fig:blockedunlockpickup}, \ref{fig:keycorridor}, and \ref{fig:obstructedmaze} respectively.

\section{Min-max ordered curricula}\label{app:ordered-curricula}

In this section, we present the corresponding min-max ordered curricula given as input to the MR\ algorithm.

For every task \(c\), we set \(m_c^1\) to \(0\) and \(M_c^1\) to \(0.5\). The real maximum mean return is around \(0.9\).

\begin{figure}[H]
\centering
\includegraphics[width=0.67\linewidth]{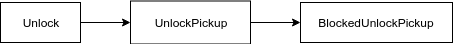}
\caption{Oriented curriculum for BlockedUnlockPickup.}
\end{figure}

\begin{figure}[H]
\centering
\includegraphics[width=\linewidth]{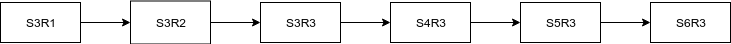}
\caption{Oriented curriculum for KeyCorridor.}
\end{figure}

\begin{figure}[H]
\centering
\includegraphics[width=.6\linewidth]{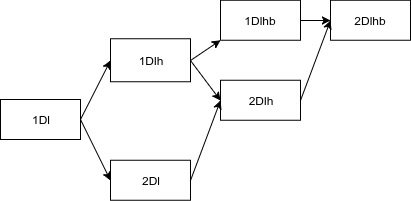}
\caption{Oriented curriculum for ObstructedMaze.}
\end{figure}

\section{Additional experimental results}

\subsection{Shortcomings of Teacher-Student algorithms}
\label{appendix:shortocmings}

\begin{figure}[H]
\centering
\includegraphics[width=.6\linewidth]{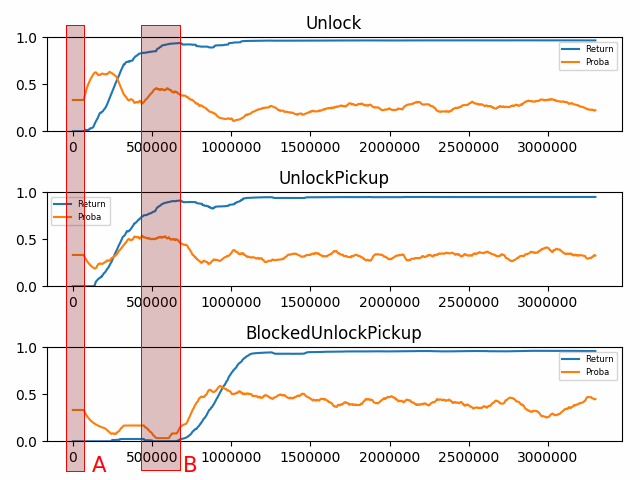}
\caption{Learning curves and task probabilities of a PPO agent trained on the BlockedUnlockPickup curriculum. Two particular moments of the training are framed.}
\label{fig:BUP-ReturnProba}
\end{figure}

We trained an RL agent on the BlockedUnlockPickup curriculum using the Teacher-Student algorithm with the Linreg attention program and the gProp distribution converter. The learning curves, along with the sampling probabilities of each task are depicted in figure \ref{fig:BUP-ReturnProba}.

Frame A of figure \ref{fig:BUP-ReturnProba} shows that, after the first few thousands time-steps, the agent has not started learning Unlock yet, because it still gets 0 reward in this tasks. However, it is trained 66\% of the time on the UnlockPickup and BlockedUnlockPickup tasks, which are significantly harder.\ Hence, it is mostly trained on tasks it can't learn yet.

Frame B of figure \ref{fig:BUP-ReturnProba} shows that, around time-step 500k, the agent already learned Unlock and UnlockPickup but is still trained 90\% of the time on them, i.e. on tasks it already learned.

\subsection{Performance of the MR algorithm on BlockedUnlockPickup}
\label{app:blockedunlockpickup}

\begin{figure}[H]
\centering
\includegraphics[width=0.5\linewidth]{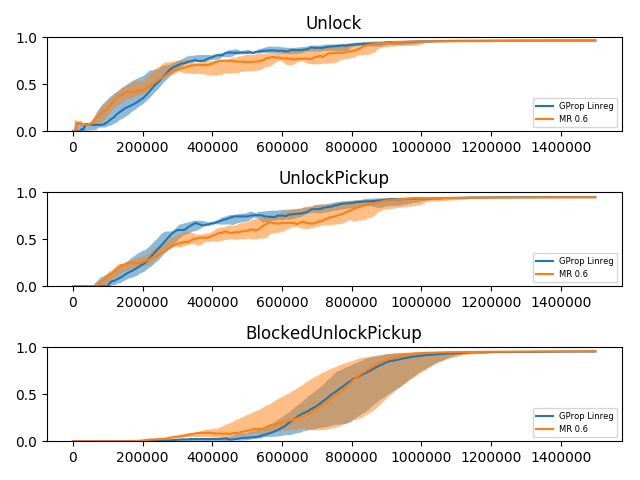}
\caption{gProp Linreg and MR with \(\delta = 0.6\) were each tested with 10 different agents (seeds) on the BlockedUnlockPickup curriculum.\  The median return during training, between the first and last quartile, are plotted.}
\label{fig:bupmr}
\end{figure}

\section{Hyperparameters}
\label{app:hyperparam}
\paragraph{Minigrid curricula}
Tables \ref{table1} and \ref{table2} summarize the hyperparameters used for all our Minigrid curricula.
\begin{table}[H]
\centering
\begin{tabular}{|c|c|}
   \hline
   $\alpha$ & 0.1 \\
   \hline
   $\varepsilon$ & 0.1 \\
   \hline
   $K$ & 10 \\
   \hline
\end{tabular}
\caption{Hyperparameters used for Teacher-Student and gProp Linreg algorithm algorithms. They are the same than those used in the Teacher-Student paper.}
\label{table1}
\end{table}

\begin{table}[H]
\centering
\begin{tabular}{|c|c|}
   \hline
   $\delta$ & 0.6 \\
   \hline
   $\gamma_{pred}$ & 0.2 \\
   \hline
   $\gamma_{succ}$ & 0.05 \\
   \hline
   $K$ & 10 \\
   \hline
   $p$ & 6 \\
   \hline
\end{tabular}
\caption{Hyperparameters used for the MR\ algorithm. }
\label{table2}
\end{table}

\paragraph{Decimal number addition curriculum} 
For the decimal number addition LSTM, we used the Adam optimizer~\citep{kingma2014adam} with a learning rate of $0.001$ for all the trained models. The batch size for the Teacher-Student algorithms was $1024$. For the MR algo, we used $128$ as a batch size, except for the Linreg + GAmax model where we had better results with a batch size of $64$. We used epochs of 10 batches in all cases. In all experiments, we used a window size $K=10$. $\delta=0.6, \gamma_{pred}=0.2, \gamma_{succ}=0.05, p=6$ were used for all MR models. $\alpha = 0.1$ was used for all models using the Window or the Online attention programs. $\varepsilon=0.1$ was used whenever we used the GAmax or the GProp A2D converters.

We tried different values for $\delta, \gamma_{pred}, \gamma_{succ}, K, p$ for the MR algorithm, and observed that the previous values were the one leading to the best performances.

\end{document}